# EEG-ITNet: An Explainable Inception Temporal Convolutional Network for Motor Imagery Classification

Abbas Salami[1], Student Member, IEEE, Javier Andreu-Perez[1,2,3,4], Senior Member, IEEE, and Helge Gillmeister[2,3]

[1]School of Computer Science and Electronic Engineering, University of Essex, Colchester, CO4 3SQ UK
[2]Department of Psychology, University of Essex, Colchester, CO4 3SQ UK
[3]The Smart Health Technologies Group, Centre for Computational Intelligence, Institute of Public Health and Wellbeing, University of Essex, Colchester, CO4 3SQ UK
[4]Simbad2, Department of Computer Science, University of Jaén, 23071 Jaén, Spain

Corresponding author: Javier Andreu-Perez (e-mail: javier.andreu@essex.ac.uk).

**ABSTRACT** In recent years, neural networks and especially deep architectures have received substantial attention for EEG signal analysis in the field of brain-computer interfaces (BCIs). In this ongoing research area, the end-to-end models are more favoured than traditional approaches requiring signal transformation pre-classification. They can eliminate the need for prior information from experts and the extraction of handcrafted features. However, although several deep learning algorithms have been already proposed in the literature, achieving high accuracies for classifying motor movements or mental tasks, they often face a lack of interpretability and therefore are not quite favoured by the neuroscience community. The reasons behind this issue can be the high number of parameters and the sensitivity of deep neural networks to capture tiny yet unrelated discriminative features. We propose an end-to-end deep learning architecture called EEG-ITNet and a more comprehensible method to visualise the network learned patterns. Using inception modules and causal convolutions with dilation, our model can extract rich spectral, spatial, and temporal information from multi-channel EEG signals with less complexity (in terms of the number of trainable parameters) than other existing end-to-end architectures, such as EEG-Inception and EEG-TCNet. By an exhaustive evaluation on dataset 2a from BCI competition IV and OpenBMI motor imagery dataset, EEG-ITNet shows up to 5.9% improvement in the classification accuracy in different scenarios with statistical significance compared to its competitors. We also comprehensively explain and support the validity of network illustration from a neuroscientific perspective. We have also made our code freely accessible at https://github.com/AbbasSalami/EEG-ITNet.

## I. INTRODUCTION

**B**RAIN-COMPUTER interfaces (BCIs) were developed to provide direct communication between the human brain and external devices through the acquisition of neural signals and their translation and transmission as control commands to assistive devices [1]. Although initially designed to help people with cognitive or physical impairments, BCIs have a wide range of applications nowadays, ranging from medical and neuroergonomics to entertainment, intelligent environments, and even security and authentication [2]. From different BCI systems based on varying neuroimaging modalities such as functional near-infrared spectroscopy (fNIRS), functional magnetic resonance imaging (fMRI), magnetoencephalography (MEG), and positron emission tomography (PET), electroencephalography (EEG) is of great importance and one of the popular techniques due to its convenient and inexpensive nature. The electrical activities of a population of neurons can generate adequately high electrical fields that can be measured by electrodes placed on the surface of the scalp. However, the recorded signals have a low signal-to-noise ratio (SNR) due to several factors such as interference from other physiological signals, electrode detachment, or signal distortion resulting from the cortex's behaviour as a volume conductor [3]. Therefore, advanced signal processing and machine learning algorithms have been extensively applied in this area to increase the SNR and the quality of the extracted neural patterns of interest [4].

Notably, we can recall many feature extraction algorithms that were applied successfully to classify motor imagery movements, including common spatial pattern (CSP) and its variants, Riemannian approaches, Fourier transform, wavelet transforms, and autoregressive models followed by different classification algorithms, such as linear discriminant analysis (LDA), support vector machine (SVM), multilayer perceptron (MLP), fuzzy systems [5-7], and k-nearest neighbors (KNN) [8, 9]. Moreover, deep learning algorithms have recently accompanied these state-of-the-art algorithms to obtain even more promising and reliable outcomes [4, 10-12]. For instance, Sakhavi et al. [13] proposed a novel temporal representation of a multi-channel EEG signal based on Hilbert transform and a convolutional neural network (CNN) to classify motor imagery movements. Wang et al. [14] trained CNN and long short-term memory (LSTM) models using multi-dimensional features derived from the representation of EEG signal using their short time Fourier transform (STFT). Although the performances of such algorithms have been superior to the conventional ones, end-to-end deep learning architectures that eliminate the need for prior information are more favoured than traditional approaches that require the extraction of handcrafted features or signals transformation pre classification. A powerful *end-to-end neural network* that permits the *explainability* of its inference can significantly impact the development of robust BCI systems and reduce the need for expertise. Hence, this paper proposes a CNN structure called EEG-ITNet based on inception modules and causal convolutional layers with dilation in the form of residual blocks.

Convolution with dilation is the foundation of temporal convolutional networks (TCNs). The TCN term was first coined by Lea et al. [15] and used as a new and effective class of temporal models for action segmentation and detection. Since then, TCNs have been suggested as a promising alternative to RNNs for sequence modelling by providing a unified approach to encode spatial-temporal information and capture high-level temporal information hierarchically. With new ways of network visualisation for interpretation, we will demonstrate how inception modules along with convolutional layers with dilation can extract rich and meaningful spectral, spatial, and temporal features from multi-channel EEG signals. The **main highlights of this paper** are:

- The proposed deep architecture has few trainable parameters, allowing its transformation into intuitive visualisation structures such as topographic maps (topo-maps). We provide a methodology for achieving this.
- A notable gain has been achieved for signal classification on the motor imagery BCI dataset in various scenarios compared to other end-to-end architectures.
- The conventional vanilla convolutions have been upgraded to inception modules and depthwise causal convolutions with dilation to achieve even higher performance with less complexity compared to similar deep learning models.

The rest of the paper will be structured as follows. In section II, we review the existing similar end-to-end neural network architectures for EEG analysis, particularly for motor imagery BCI systems. Then in section III, we extensively describe our method and architecture by explaining the fundamentals first. The section also contains full details of our visualisation techniques. In section IV, we confirm the effectiveness of our proposed network on EEG signal classification in three different scenarios followed by network visualisation. Finally, we discuss and conclude the results and extracted features in sections V and VI, respectively, including our future work plan.

## II. RELATED WORK

In the literature of end-to-end architectures, Schirrmeister et al. [16] proposed ShallowNet and DeepConvNet for EEG decoding based on a raw dataset, which consisted of a shallow and deep sequence of convolutional layers, respectively. Although they reported comparable results with the state-of-the-art methods, the high number of parameters, especially in DeepConvNet, made the networks too complex to analyse and interpret. Later, Lawhern et al. [17] introduced EEGNet, a more compact and efficient CNN architecture with few parameters and fast training nature. EEGNet showed encouraging results on several types of EEG datasets with significantly fewer parameters than ShallowNet and DeepConvNet. In addition, the simple architecture of EEGNet has made it a notable candidate for EEG analysis in different scenarios. Although the authors attempted to visualise and interpret the features extracted from the network, the outcome is not very intuitive. In a more recent study, EEG-Inception [18] was introduced as an end-to-end architecture utilising inception modules originally proposed by Szegedy et al. [19] in computer vision. However, although EEG-Inception reached superior results than other CNN architectures, the network lacked interpretability as the authors also made no attempt in explaining the learning process and extracted features. Similarly, we should also recall MI-EEGNET [20] developed based on Inception and Xception architectures for motor imagery classification. Although MI-EEGNET performed better than previous architectures, the high number of trainable parameters made it challenging to interpret. Besides these CNN architectures investigated for EEG signal classification, several studies can be found in the literature that used recurrent neural network (RNN) and its variants [21-23], such as LSTM and gated recurrent units (GRU), for the classification of mental tasks based on EEG signals [24, 25]. However, RNNs are less prevalent in this area due to their exploding/vanishing gradient or lack of memory problems [26].

TABLE I
COMPARISON OF THE KEY FACTORS OF EEG-ITNET WITH OTHER END-TO-END ARCHITECTURES

| Property \ Work | Inception | Dilation | Visualisation | Interpretability | Deepness (# layers) | Parametric complexity |
|---|---|---|---|---|---|---|
| **EEGNet 8,2 [14]** | No | No | Yes | Partial | Low (7) | Low (~2k) |
| **EEG-TCNet [23]** | No | Yes | No | No | Medium (17) | High (~5k) |
| **EEG-Inception [15]** | Yes | No | No | No | High (23) | Very high (~15k) |
| **EEG-ITNet** | Yes | Yes | Yes | Yes | Very high (31) | Medium (~3k) |

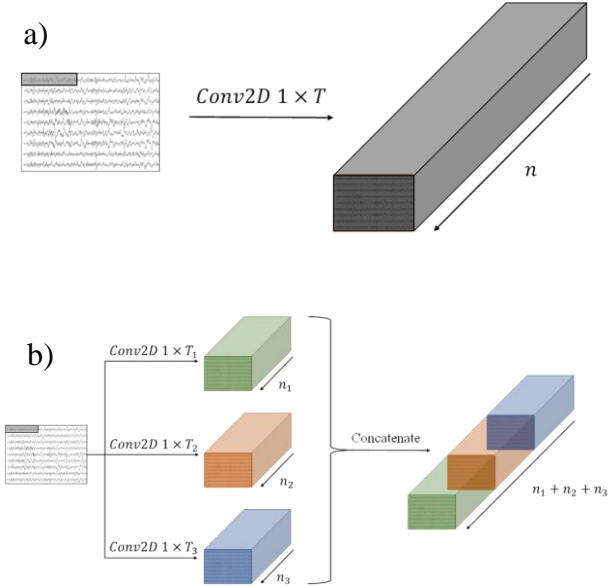

Fig. 1. Conventional convolutional layer (a) and its alternative with inception modules (b)

In response to the slow training essence of RNNs for EEG signal analysis, TCNs have been recently applied and demonstrated promising results for temporal analysis of EEG time series with faster computation. In this regard, Ingolfsson et al. [26] and Musallam et al. [27] have reported superior results for motor imagery signal classification by adding TCN in their structure. TCNs have also been successfully applied before to analyse different EEG datasets, including mental workload assessment [28], pathological versus non-pathological EEG classification [29], and recently for quantitative scoring of depression [30]. Table I compares the key factors of EEG-ITNet with some other existing end-to-end neural networks, which we later use as comparison models.

## III. METHOD

This section extensively explains the fundamental concepts and blocks of EEG-ITNet, followed by a detailed illustration of the proposed architecture. This section also contains a description of our visualisation and interpretation process.

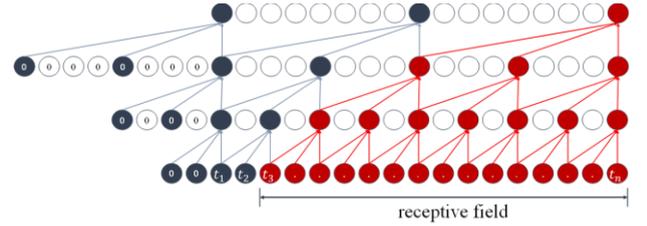

Fig. 2. Three causal convolutional layers with leading zero padding and the corresponding receptive field. In this example, the kernel size is 3 and the dilation based is 2.

### A. BACKGROUND

This subsection defines the theories used to form the fundamental blocks of EEG-ITNet.

*1) Inception:* In order to break the multi-channel EEG signal into its informative frequency sub-bands, this study uses several parallel (rather than sequential) convolutional layers with various filter sizes and stacks their output along the convolutional channel dimension (Fig. 1-b), rather than adopting a single convolutional layer with a fixed-length kernel (Fig. 1-a). This approach is based on the idea of the naive inception module, initially proposed in computer vision [19] to tackle the drawbacks of using deeper, wider, and more complex networks for image classification. Inception modules are effective, especially in the case of small training datasets or limited computational capacity, where it is not feasible to use a deep, complex neural network with a high number of parameters. This is because the network gets either more prone to overfitting or goes without the computational resources [19].

*2) Dilation:* Adding dilation to a series of convolutional layers can increase the receptive field of the network. Therefore, we applied this modification of CNN consisting of a series of causal convolutional layers with varying dilation (powers of a base dilation) to take advantage of a more extended history coverage and extract rich temporal features at each time step. By causal convolutional, we mean that the output of the convolutional layer at each time step depends only on the earlier time steps of the input time series (Fig. 2). A leading zero padding has been applied to achieve this and ensure an output sequence with equal length to the input sequence. Using CNN with dilation is the basis of TCNs, which were initially proposed by Bai et al. [31] as a valuable alternative tool to RNNs for sequence modelling. Mathematically, if we have $n$ convolutional layers stacked on one another with a dilation base of $b$ and a kernel size of $T$, then the TCN receptive field $r$ can be calculated as follows:

$$r = 1 + (T-1) \cdot \frac{b^n - 1}{b - 1} \quad (1)$$

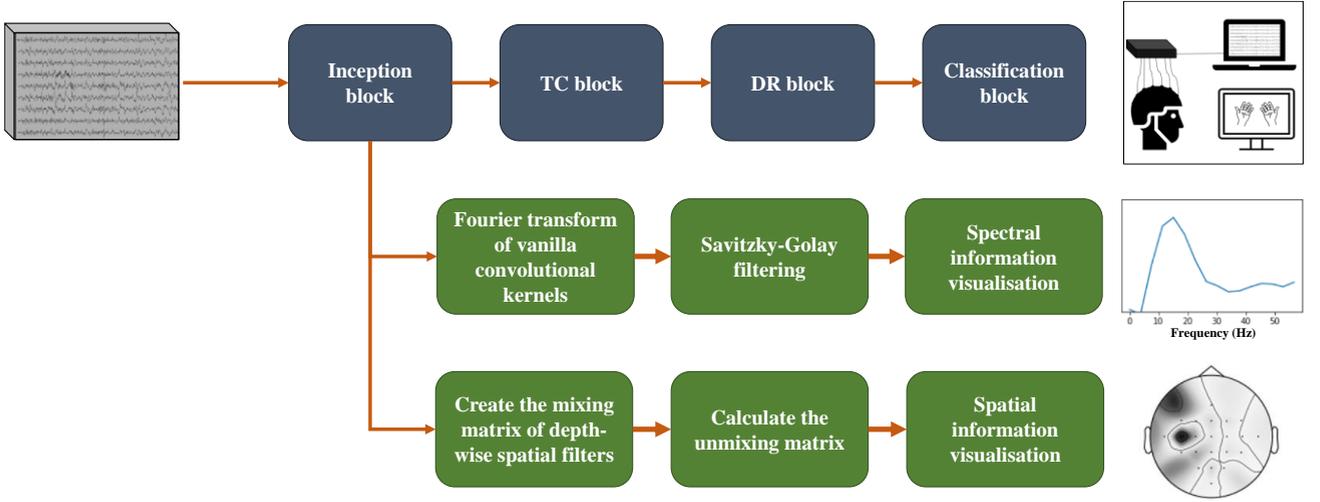

Fig. 3. General schematic of EEG-ITNet

Notice that the kernel size $T$ needs to be selected as a number greater than the dilation base $b$ to avoid having holes in the receptive field. By holes, we mean samples in the input sequence that do not affect the output sequence. To further enhance the performance of TCN, its representation in the form of residual blocks was proposed [31]. For the extraction of temporal features, we designed a TCN as a series of residual blocks, each consisting of multiple causal convolutional layers with the same dilation. The dilation rate increases for each consecutive residual block as the powers of the dilation base. In this case, with having $n$ residual blocks with $m$ convolutional layers in each, a dilation base of $b$, and a kernel size of $T$, the receptive field of the TCN will be calculated as follows:

$$r = 1 + m(T-1) \cdot \frac{b^n - 1}{b - 1} \quad (2)$$

So (2) can be used to determine the number of residual blocks and causal convolutions in each, dilation base, and the kernel size needed to design a TCN that can account for $r$ time steps back in time. Of course, the residual blocks also contain activation functions and dropout to account for nonlinearity and avoid overfitting in the network, respectively. We also applied batch normalisation after convolutional layers to tackle the problem of exploding gradient.

### B. PROPOSED DEEP ARCHITECTURE
The general architecture of EEG-ITNet is depicted in Fig. 3 and consists of 4 main blocks: inception block, temporal convolution (TC) block, dimension reduction (DR) block, and classification block.

*1) Inception block:* The learning process starts with three parallel sets of layers; each includes a 2D convolutional layer along the time axis, which acts as frequency filtering, followed by a 2D depthwise convolutional layer acting as spatial filtering. Adding inception modules with varying convolutional kernel sizes eliminates the need for a fixed-length kernel [18] and allows the network to learn filters representing various frequency sub-bands. We will later show that the longer the kernel size, the more likely to learn features in low-frequency components. The kernel size for the depthwise convolution equals the number of electrodes in the dataset to design a spatial filter that combines all the electrodes to find the sources of brain activity. Hence, the tensor obtained after the inception modules represent the signals of sources in different frequency sub-bands. This block ends with a nonlinear activation function and dropout to allow the network to learn more complex nonlinear spatial information and avoid overfitting, respectively.

*2) Temporal convolution (TC) block:* After extracting sources in different informative frequency sub-bands, the TCN architecture is applied to extract the discriminative temporal features while taking the history of the time series into account. The TC block consists of several residual blocks, and each is formed by depthwise causal convolutional layers with leading zero padding, followed by activation function and dropout. Using depthwise causal convolution followed by batch normalisation instead of weight normalisation is a modification that we made in the original TCN structure. Since the output of the inception block represents signals in the source domain, depthwise causal convolution has been used to ensure that the temporal information of each source is extracted separately. With this modification, our observation showed more robust and superior performance than the conventional TCN. Depending on the number of previous time steps intended to be considered, the number of residual blocks, kernel length, and dilation base can be chosen based on (2). An average pooling layer also precedes this block to reduce the data dimensions and avoid overfitting.

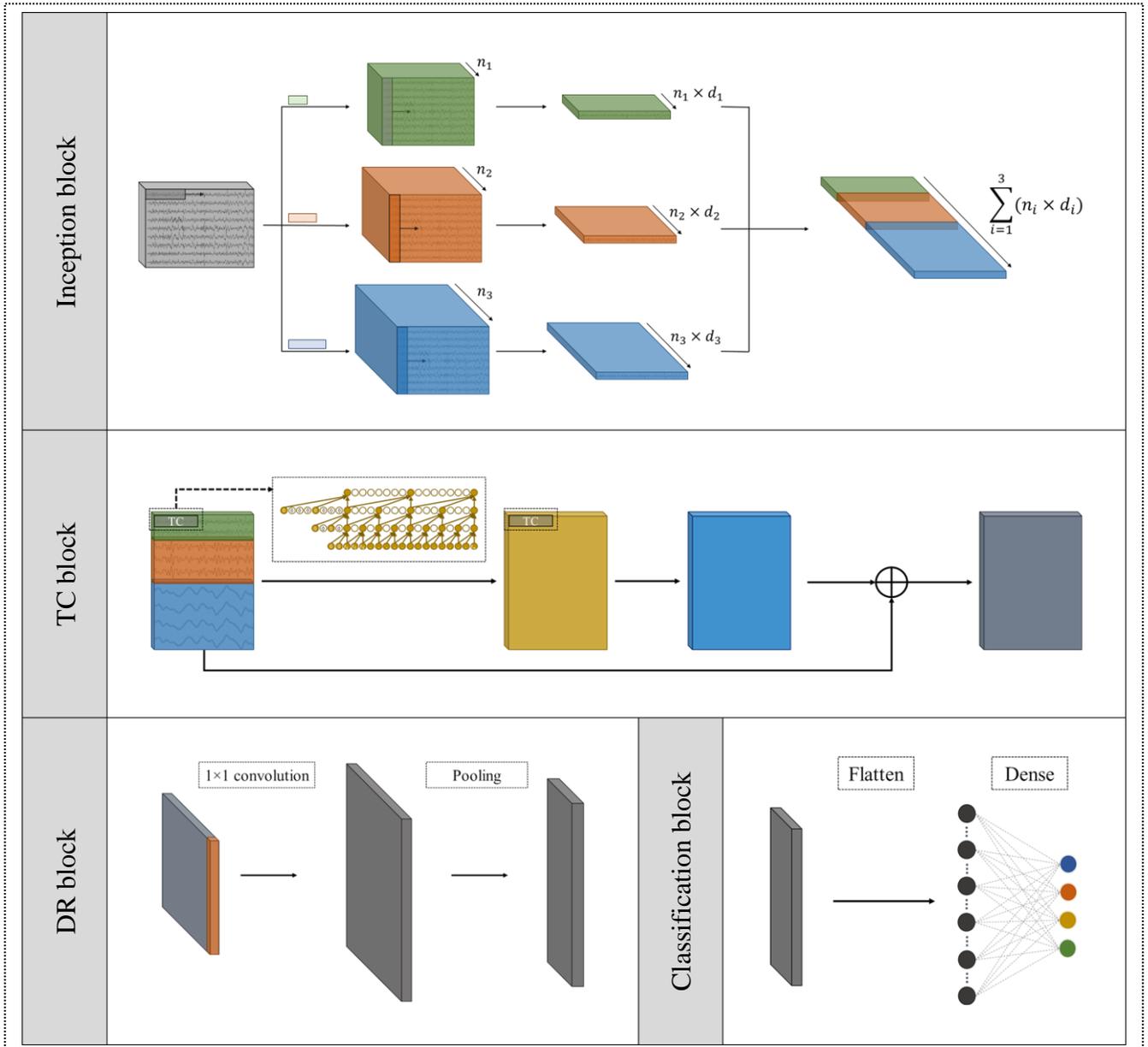

Fig. 4. Details of different blocks in EEG-ITNet architecture

*3) Dimension reduction (DR) block:* The output of the TC block essentially contains temporal information extracted from sources with various frequency spectrums. Thus, we utilised a $1 \times 1$ convolutional layer to combine these temporal features and control the number of final features used to perform the classification task. The number of convolutional filters is a hyperparameter and needs to be adjusted to avoid overfitting. As well as an activation function and a dropout layer, this block also ends with an average pooling layer to further reduce the tensor dimension.

*4) Classification block:* This block is the final piece of EEG-ITNet and contains a fully connected layer with a "softmax" activation function that follows a flatten layer. Although we call it the classification layer, it can be easily modified depending on the desired output and the problem set.

### C. INTERPRETABILITY AND VISUALISATION OF EEG-ITNet

As depicted in Fig. 4, the inception block consists of 2D convolutional layers applied over the time axis of multi-channel EEG with varying kernel sizes. Adding an extra dimension to the input signal preserved for convolutional channels, these initial 2D convolutional layers act as frequency filters and extract the signal in different informative sub-bands. To demonstrate this, assume the input signal in each electrode $X[n] \in R^{1 \times s}$, where $s$ is the number of samples. Since a convolutional layer is simply the dot product of the input signal and a kernel, the output $X'[n] \in R^{1 \times s}$ of the initial vanilla 2D convolutional layer with a kernel $K$ of size $2l + 1$ and "same" padding over each EEG channel can be calculated as follows:

$$X'[n] = \sum_{i=-l}^{l} X[i]K[i-n] \qquad (3)$$

which represents the convolution between the EEG time series of each electrode and the reverse of the kernel ($X[n] * K[-n]$). We also know that the convolution in the time domain acts as multiplication in the frequency domain. Therefore, taking the Fourier transform of convolutional kernels, we can find out the frequency sub-bands selected by the network during the training phase. However, since the deep learning algorithms are sensitive to even tiny discriminative features, the Fourier transform of convolutional kernels can be highly varied. So for the sake of visualisation, we use Savitzky-Golay filtering [32] to smooth the Fourier transforms. Savitzky-Golay is a type of digital filtering consisting of a series of least-square polynomial approximations applied on fixed-length time windows swept over the time series. For a sequence of samples $Y[n]$, the mean-squared approximation error for a time window centred at $c$ can be calculated as follows:

$$E = \sum_{n=c-l}^{c+l} (\sum_{m=0}^{p} a_m n^m - Y[n])^2 \qquad (4)$$

where $a_m$, $m = 0, ..., p$ are polynomial coefficients, $p$ is the polynomial order, and $l$ is the half-width of the filter window. To minimise the mean-squared approximation error and find the optimal polynomial coefficients, we should take the derivative of (4) with respect to all the polynomial coefficients and set them equal to 0, which yields a set of $p + 1$ normal equations [33]. Accordingly, the polynomial coefficients can be found as follows:

$$A = (D^T D)^{-1} D^T Y \qquad (5)$$

where $A = [a_0, a_1, ..., a_m, ..., a_p]^T$ is the vector of polynomial coefficients. The matrix $D = \{d_{n,m}\}$ is called the design matrix with the size of $2l + 1$ by $p + 1$ and the elements as follows:

$$d_{n,m} = n^m, \quad \begin{array}{l} c - l \leq n \leq c + l \\ m = 0, ..., p \end{array} \qquad (6)$$

The Savitzky-Golay smoothing process can also be explained as a shift-invariant discrete convolution process, which is why it is referred to as Savitzky-Golay filtering (details can be found in [33]).

Besides batch normalisation and activation function, each inception module in EEG-ITNet also contains a depthwise convolutional layer. The depthwise convolutional layer with "valid" padding acts as spatial filtering. It linearly combines signals in different electrodes to transform the surface domain to the source domain and find discriminative sources. Note that this convolutional layer is applied separately in each convolutional channel, as each convolutional channel represents the signal in a different frequency sub-band. Also, the nonlinear activation function that follows this layer further improves the spatial filtering by giving it nonlinearity. Thus, for an input EEG signal $X[n] \in R^{c \times s}$ with $c$ channels and $s$ number of samples in each channel, the transformation from the surface domain to the source domain $S[n] \in R^{c' \times s}$, with $c'$ being the number of sources, can be formulated as follows:

$$S[n] = WX[n] \qquad (7)$$

where $W$ is a $c' \times c$ matrix containing spatial filters; however, $W$ in (7) represents the unmixing matrix and does not represent the spatial locations of sources. Therefore, we must find and visualise the mixing matrix $W^{-1}$ to find the correct location of sources learned during the training phase.

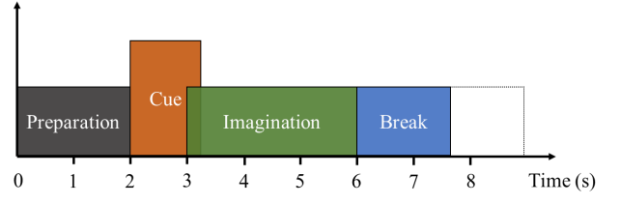

Fig. 5. Timing scheme of the motor imagery dataset 2a from BCI competition IV

## IV. EXPERIMENTS AND RESULTS

### A. DATASET AND EVALUATION SCENARIOS

Dataset 2a of BCI competition IV and OpenBMI motor imagery dataset [34] were used in this study to evaluate the performance of EEG-ITNet exhaustively. The description of each dataset is as follows:

*1) BCI competition IV dataset 2a:* It is a multi-class motor imagery dataset containing EEG recordings of 9 participants during imagination of left hand, right hand, both feet, and tongue movements. The dataset has been collected in two separate sessions, and each consists of 288 trials with an equal number of trials for each class. The actual labels were provided for the data in the first session, while the second session was reserved for testing the classification algorithms in the competition. Each trial in the paradigm started with 2 seconds of preparation time, followed by a cue that lasted for 1.25s, representing the imagination class (Fig. 5). The imagination period continued for 4s after the cue onset and was terminated by an inter-trial break. The EEG signals were recorded using 22 electrodes and sampled with 250Hz. Since we aimed to propose an end-to-end neural network, the signals were only downsampled to 125Hz, scaled to have zero mean and unit variance, and epoched to a 3-seconds time window after the cue onset. It should also be noted that the 3 EOG channels available in the dataset were excluded from our analysis.

*2) OpenBMI motor imagery dataset:* Collected by Lee et al. [34], OpenBMI is a large EEG dataset containing EEG recordings from 54 subjects across multiple sessions performing three different BCI tasks: motor imagery (MI), event-related potential (ERP), and steady-state visually evoked potential (SSVEP). In this study, we used EEG recordings from the first session of the MI task, which represent neural activities while performing the imagery task of grasping with either left or right hand. The experiment consists of balanced train and test datasets with 100 trials each. The experimental paradigm of this dataset is depicted in Fig. 6. It starts with 3s of preparation period followed by 4s of imagination period based on the movement direction of the fixation cross. 62 Ag/AgCl electrodes were used to record the signals at 1kHz sampling frequency. Similar to BCI competition IV dataset 2a, the dataset was only downsampled to 125Hz, standardised to have zero mean and unit variance, and epoched to a 3-seconds time window after the cue onset. We should note that similar to what the authors proposed in the original paper [34], only 20 electrodes located on the motor cortex (FC-5/3/1/2/4/6, C5/3/1/z/2/4/6, and CP-5/3/1/z/2/4/6) were selected for the analysis.

The performance of EEG-ITNet was evaluated in three different scenarios, corresponding to within-subject, cross-subject, and cross-subject with fine-tuning, detailed below. Note that, since the data from the second sessions were used for testing the system and the labels were unknown to the participants in the competition, we did not use them as training or validation sets in our analysis.

*1) Within-subject:* For each subject independently, the EEG recordings of their first session were divided into train and validation set to find the best parameters of the network. Then, the trained network was evaluated on the EEG recordings of the test session, and the results were reported for each subject.

*2) Cross-subject:* The network parameters were learned using the training set of all other subjects rather than the test subjects themselves, meaning the system had never seen any data from the test subject. The subjects preserved for training were divided to form our train and validation sets. Similar to the within-subject case, unique network parameters were obtained for each individual subject in this case.

*3) Cross-subject with fine-tuning:* In the final scenario, the labelled EEG recordings of the training session of each target subject were used to fine-tune the parameters of the system developed for each target subject in the second scenario.

### B. COMPARISON MODELS

We have selected three similar existing end-to-end architectures for an exhaustive evaluation of the EEG-ITNet performance, including EEGNet, EEG-TCNet, and EEG-Inception. This section explains each of these networks and their implementation in this study.

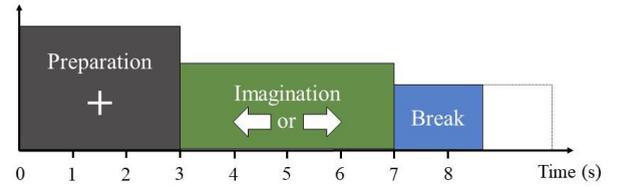

Fig. 6. Timing scheme of the OpenBMI motor imagery dataset

*1) EEGNet:* As a compact convolutional neural network, EEGNet showed promising results for analysing different EEG datasets [17]. Like DeepConvNet, EEGNet starts with temporal and spatial convolutions, followed by a separable convolution.

*1) EEG-TCNet:* Proposed as an extension of EEGNet, EEG-TCNet applied a TCN structure after feature extraction layers of EEGNet to further improve its performance [26].

*3) EEG-Inception:* Designed as a classification tool for ERP-based spellers, EEG-Inception is a CNN architecture that comprises two inception modules and an output module [18]. Each inception module consists of three parallel sets of layers, including 2D vanilla convolutions.

Since our dataset and preprocessing stages are similar to Lawhern et al. [17], we used the same implementation of EEGNet. The EEG-Inception was also implemented based on table IV in [18] as they used a dataset with the same sampling frequency as ours (after downsampling). For EEG-TCNet, the authors introduced their network as an extension to the shallow EEGNet. So in order to reproduce EEG-TCNet to be compatible with our preprocessing stages, we used the same implementation of EEGNet provided in [17] followed by a TCN structure. The parameters of TCN were selected based on the first formula in [26] as follows: $K_T = 3, L = 2, F_T = 16, p_t = 0.3$.

### C. TRAINING PROCESS AND HYPERPARAMETERS SELECTION

The inception block in EEG-ITNet consists of three 2D convolutional layers with 2, 4, and 8 filters and kernel sizes of 16, 32, and 64 samples, respectively, followed by three depthwise convolutional layers with a depth of 1. Since the input signal's sampling rate is 125Hz, the largest kernel was selected with the size of 64 samples as it can capture frequency components as low as $125/64 \approx 2Hz$ of the input signal.

For example, Fig. 10 contains the learned frequency filters with the 16, 32, and 64-sample kernels for subject 3 in BCI competition IV dataset 2a. As it can be seen, the larger the filter size, the more it is prone to capture lower frequencies. The reason behind selecting more filters for larger kernel sizes is the fact that motor imagery is often associated with activities in the lower alpha (8–12Hz) and beta (13–30Hz) bands [35], rather than high frequency bands, e.g. gamma (30–100 Hz) band. Besides, the lack of samples in short filter sizes makes it difficult to calculate their Fourier transform and visualise them. Notwithstanding, gamma activity can still be observed

TABLE II
PERFORMANCE EVALUATION FOR WITHIN-SUBJECT CASE IN TERMS OF CLASSIFICATION ACCURACY FOR BCI COMPETITION IV DATASET 2A

|         | EEG-Inception | EEGNet 8,2   | EEG-TCNet    | EEG-ITNet    |
|---------|---------------|--------------|--------------|--------------|
| S1      | 77.43 (4th)   | 81.94 (3rd)  | 82.29 (2nd)  | **84.38 (1st)** |
| S2      | 54.51 (4th)   | 56.94 (3rd)  | **64.24 (1st)** | 62.85 (2nd)  |
| S3      | 82.99 (4th)   | **90.62 (1st)** | 88.89 (3rd)  | 89.93 (2nd)  |
| S4      | **72.22 (1st)** | 67.01 (3rd)  | 60.76 (4th)  | 69.1 (2nd)   |
| S5      | 73.26 (2nd)   | 72.57 (4th)  | 72.92 (3rd)  | **74.31 (1st)** |
| S6      | **64.24 (1st)** | 58.68 (3rd)  | 62.5 (2nd)   | 57.64 (4th)  |
| S7      | 82.64 (3rd)   | 76.04 (4th)  | 83.33 (2nd)  | **88.54 (1st)** |
| S8      | 77.78 (4th)   | 81.25 (2nd)  | 79.51 (3rd)  | **83.68 (1st)** |
| S9      | 76.39 (3rd)   | 78.12 (2nd)  | 76.39 (3rd)  | **80.21 (1st)** |
| Average | 73.50         | 73.69        | 74.54        | **76.74**    |
| Std     | 9.11          | 11.12        | 10.09        | 11.48        |
| p-value | **0.043*** | **0.010*** | 0.055        | -            |

*Significant at level of 0.05

during motor imagery with open eyes due to changes in subjects' spatial attention toward the target limb [36].

The inception block is followed by an average pooling layer with a pool size of 4. Afterwards, to extract temporal features in the TC block, we decided to use four residual blocks, each consisting of two convolutional layers, with a dilation base of 2 capable of covering the whole signal history. For this purpose, the TC block requires convolutional layers with a filter size of 4, which can be calculated based on (2). Finally, the number of filters in the $1\times1$ convolutional layer used as a dimension reduction technique was selected to be 14, followed by another average pooling with a pool size of 4. The dropout rate of 0.4 and 0.2 was also picked throughout the network for within-subject and cross-subject scenarios, respectively.

In all scenarios, we first used 10-fold cross-validation with 500 (150 in cross-subject) epochs in each to select the best model parameters, with the help of early stopping with the patience of 100 (15 in cross-subject) to avoid overfitting. Then we trained the networks for a maximum of 50 extra epochs with the learning rate of $10^{-4}$ on the combined training and validation sets to benefit from the information of all the labelled samples. The number of extra epochs needs to be carefully investigated to ensure no overfitting. For this purpose, we extracted the curves representing train and test losses versus a range of extra epochs used in the training process, concatenated to the loss patterns in the selected best fold of cross-validation. For instance, Fig. 7 shows this curve for subject 3 in BCI competition IV dataset 2a while training

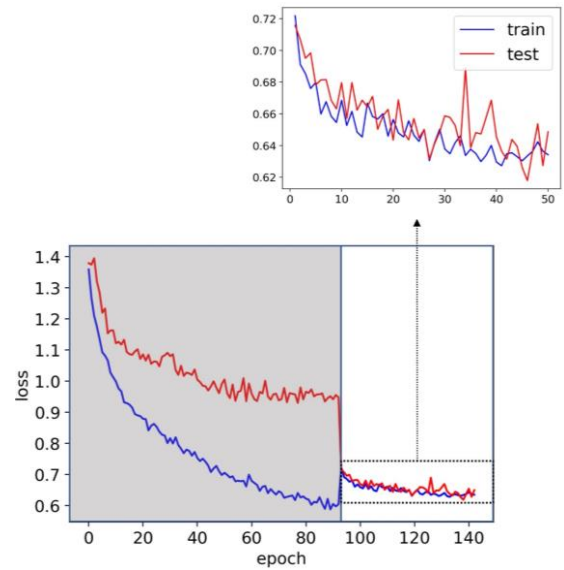

Fig. 7. The effect of the number of extra epochs in the training of EEG-ITNet for subject 3 from BCI competition IV dataset 2a in the cross-subject scenario. Left (grey box) - training process in the selected best fold of cross-validation. Right (white box) - training process for 50 extra epochs using all labelled data

EEG-ITNet in the cross-subject scenario. Notice that the sudden jump in the curves is due to the fact that in each fold, we split all labelled data to form train and validation sets, while during extra epochs, we use all labelled data as the train set, resulting in a different range of values for our losses. It

TABLE III
PERFORMANCE EVALUATION FOR CROSS-SUBJECT CASE IN TERMS OF CLASSIFICATION ACCURACY FOR BCI COMPETITION IV DATASET 2A

|  | EEG-Inception | EEGNet 8,2 | EEG-TCNet | EEG-ITNet |
|---|---|---|---|---|
| S1 | 66.32 | 68.75 | 69.1 | **71.88** |
| S2 | 48.26 | 50 | 52.08 | **62.85** |
| S3 | 73.61 | 80.21 | **81.94** | **81.94** |
| S4 | 56.6 | 59.38 | 61.81 | **65.62** |
| S5 | **65.62** | 64.24 | 60.42 | 63.19 |
| S6 | **56.25** | 48.26 | 51.39 | **56.25** |
| S7 | 73.61 | 72.57 | 76.39 | **80.21** |
| S8 | 70.49 | 77.43 | 74.31 | **78.12** |
| S9 | 61.11 | 55.56 | 58.68 | **64.93** |
| Average | 63.54 | 64.04 | 65.12 | **69.44** |
| Std | 8.69 | 11.59 | 10.86 | 8.98 |
| p-value | **0.009*** | **0.008*** | **0.006*** | - |

*Significant at level of 0.05

TABLE IV
PERFORMANCE EVALUATION FOR CROSS-SUBJECT WITH FINE-TUNING CASE IN TERMS OF CLASSIFICATION ACCURACY FOR BCI COMPETITION IV DATASET 2A

|  | EEG-Inception | EEGNet 8,2 | EEG-TCNet | EEG-ITNet |
|---|---|---|---|---|
| S1 | 77.43 | 84.38 | **86.46** | 84.03 |
| S2 | 54.86 | 54.86 | 64.93 | **65.28** |
| S3 | 87.85 | **92.36** | 90.28 | 92.01 |
| S4 | 72.57 | 67.01 | 71.53 | **73.96** |
| S5 | 74.65 | 66.67 | 73.26 | **75.35** |
| S6 | **66.32** | 61.46 | 58.68 | 64.93 |
| S7 | 79.17 | 79.86 | 80.56 | **84.72** |
| S8 | 83.33 | 82.99 | 82.99 | **84.72** |
| S9 | 79.17 | 75.69 | 73.61 | **83.68** |
| Average | 75.04 | 73.92 | 75.81 | **78.74** |
| Std | 9.76 | 12.19 | 10.24 | 9.4 |
| p-value | **0.008*** | **0.010*** | **0.022*** | - |

*Significant at level of 0.05

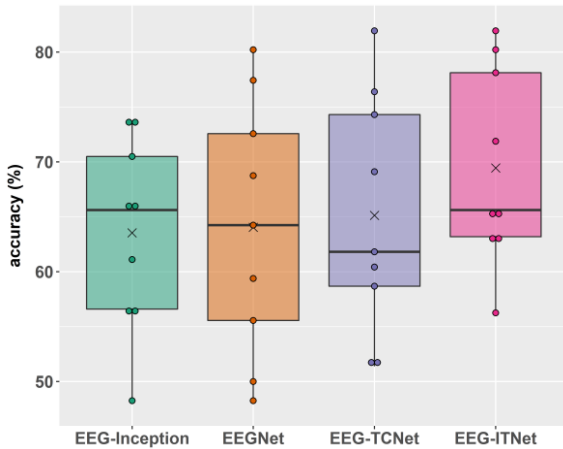

Fig. 8. Box plot of the classification performance on BCI competition IV dataset 2a in the cross-subject scenario

should be reminded that in the competition, the participants did not have access to the test data labels to evaluate the performance of their proposed algorithm. However, in this paper, for the sake of fair comparison, we selected the best test accuracy for each deep learning model and each subject, meaning the number of extra epochs varies over models and subjects.

### D. PERFORMANCE EVALUATION

Table II summarises the classification accuracies in the within-subject case for EEG-ITNet and our comparison models on BCI competition IV dataset 2a. The performance of EEG-ITNet has been superior to other classification algorithms in terms of mean classification accuracy over all the subjects. Nevertheless, a one-sided Wilcoxon signed-rank test with a significance level of 0.05 was used to verify the significance of improvement [37]. The Wilcoxon signed-rank test is a non-parametric statistical test, especially suitable for small sample sizes. Non-parametric statistical tests generally do not require large sample sizes and make fewer assumptions about the data distribution, including normality. The performance of EEG-ITNet is very promising, reaching the highest accuracy for five out of nine subjects and the second best in three cases.

Table III shows the performance of EEG-ITNet in the cross-subject scenario and its comparison with other end-to-end architectures on BCI competition IV dataset 2a. It also contains the p-value needed to reject the null hypothesis (no difference in the classification accuracies obtained by EEG-ITNet and each comparison model). In this scenario, each model is trained without using any data from the target subject. Therefore, due to the dynamic nature of EEG signals and their variation from one subject to another, significantly lower performance have been observed in this case. However, our proposed model has managed to reach the highest mean accuracy in the cross-subject case, with statistically significant improvement over EEG-Inception, EEGNet, and EEG-TCNet. Fig. 8 also shows this comparison and the data distribution of the results reported in Table III using a box plot.

Finally, for each subject, their EEG recordings from the training session were used to fine-tune the parameters of the systems designed for them in the cross-subject scenario. The classification results of this case can be found in Table IV,

TABLE V
SUMMARY OF CLASSIFICATION PERFORMANCE IN DIFFERENT SCENARIOS FOR OPENBMI MOTOR IMAGERY DATASET

| | Within-subject | | | Cross-subject | | | Cross-subject with fine tuning | | |
|---|---|---|---|---|---|---|---|---|---|
| | acc | t (n) | *p-value* | acc | t (n) | *p-value* | acc | t (n) | *p-value* |
| EEG-Inception | 69.3 | 3.88 | *<0.001*\* | 71.15 | 3.71 | *<0.001*\* | 75.11 | 1.52 | 0.067 |
| EEGNet 8,2 | 69.61 | 4.23 | *<0.001*\* | 71.2 | 3.35 | *<0.001*\* | 74.04 | 3.35 | *<0.001*\* |
| EEG-TCNet | 68.35 | 5.23 | *<0.001*\* | 70.83 | 4.55 | *<0.001*\* | 73.85 | 4.32 | *<0.001*\* |
| EEG-ITNet | **71.91** | - | - | **73.52** | - | - | **76.19** | - | - |

acc: Classification accuracy, t: Test statistic T, n: Degree of freedom
*Significant at level of 0.05

which confirm the superior performance of EEG-ITNet compared to other deep learning models with statistical significance. We have also investigated the improvement in the classification results compared to the within-subject scenario in Fig. 9. Comparing the improvement of shallow EEGNet (5 out of 9 subjects with an average improvement of 0.23%) with deep EEG-ITNet (7 out of 9 subjects with an average improvement of 2%) proves the ability of the proposed deeper architecture to handle extra information from other subjects more effective than a shallow one. The improvement is also more significant for complex networks (in terms of the number of parameters) such as EEG-Inception, which showed improvement in 7 out of 9 subjects with an average improvement of 1.54%. However, deep EEG-ITNet is more favoured than the deep and complex EEG-Inception in this area, as it has more potential to be visualised and interpreted.

Because of the high number of subjects in the OpenBMI motor imagery dataset, results have been summarised in Table V. The table shows superior performance for EEG-ITNet in all of our cases compared to other deep models in terms of mean classification accuracy. In order to evaluate the significance of improvement, we performed right-tailed paired t-test after checking the test assumptions. We confirmed data normality by Shapiro–Wilk test.

### E. VISUALISATION

Fig. 10 shows the spectral and spatial filters learned for subject 3 from BCI competition IV dataset 2a in the inception block of EEG-ITNet using the visualisation techniques proposed in section III. This has been selected as an example to show how each set of frequency response and scalp topo-map represents the power spectrum and spatial location of a group of sources activating uniquely among motor imagery classes. Notice that the sign of spatial coefficients does not carry any information about oscillatory synchronisation or desynchronisation processes. The network is trained to perform a classification

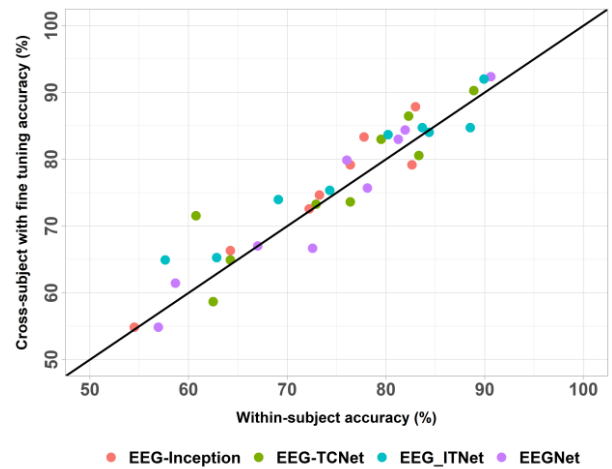

Fig. 9. Classification improvement in cross-subject with fine-tuning case compared to the within-subject scenario for different deep learning models on BCI competition IV dataset 2a task by finding discriminative neural patterns among the four motor imagery classes. The spatial activations' location and the absolute value of coefficients are important only as they represent the occurrence and significance of distinct neural activities in at least one of the motor imagery classes. So in order to avoid confusion, the spatial patterns are plotted in grayscale symmetric to zero.

As shown in Fig. 10, larger filter sizes capture relatively lower frequencies while smaller filters tend to learn higher frequency components. To explain the learned filters, we should consider the dataset to be composed of two tasks, including motor imagery and sustained spatial attention toward the target limb. Motor imagery is associated with a decrease in alpha and beta power (event-related desynchronisation) over the sensory and motor cortex in the contralateral hemisphere [35, 38, 39]. This modulated alpha activity has been clearly captured by filters 4, 5, 7, and 8 in the left panel of Fig. 10 and filter 4 in the middle panel. Notice that among these filters, spatial activation is located at the vertex only in filter 5 of the left panel, which is likely to be associated with the imagination of bilateral foot movement. In other cases, the filters represent modulated neural patterns during left or right hand motor imagery located in the contralateral hemisphere. The beta equivalent to the alpha desynchronisation seen over the vertex (filter 5 of the 64-sample kernel) during motor imagery was also captured by 64-sample filters 1 and 6, which may be similarly related to bilateral foot movement imagination.

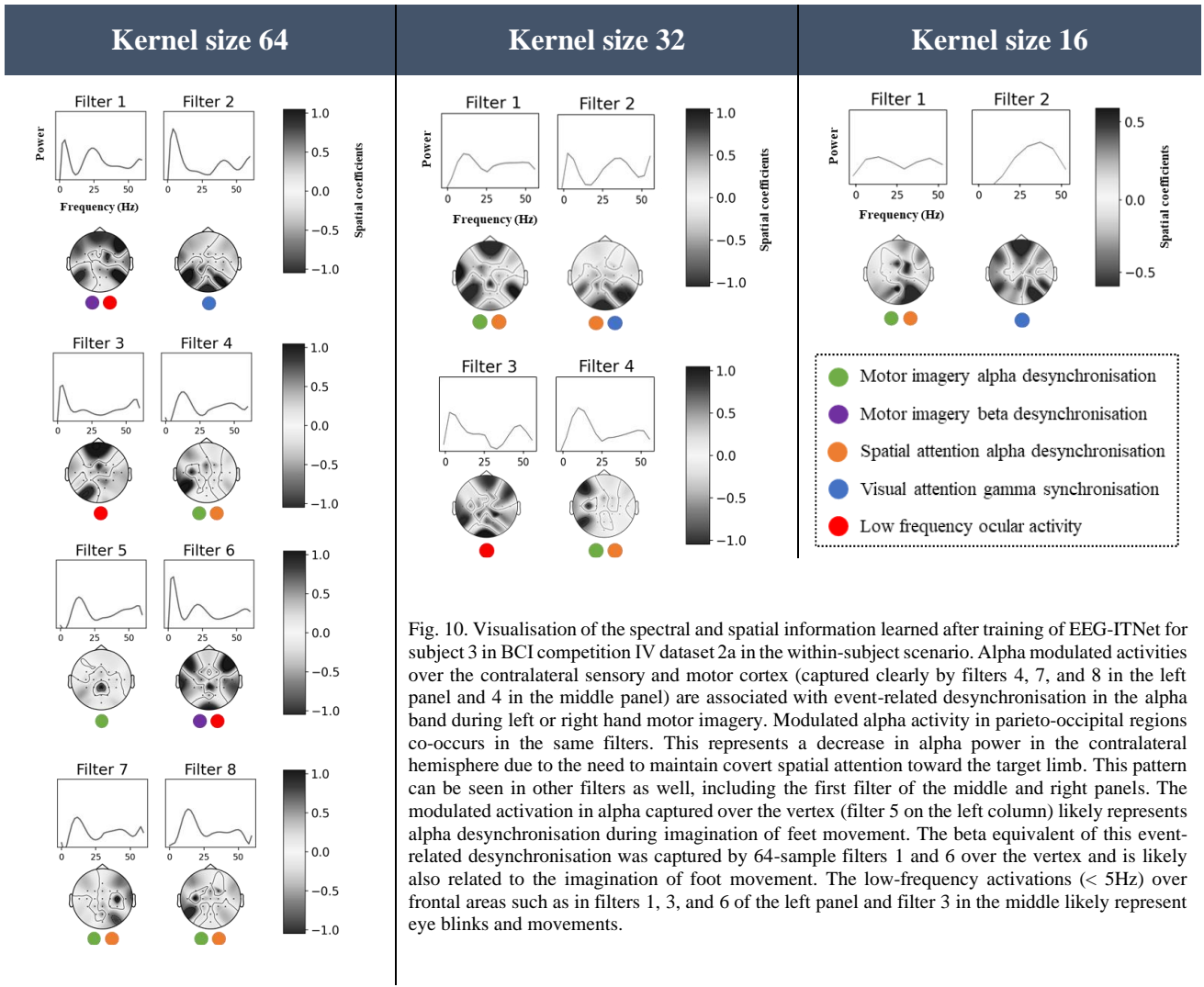

Fig. 10. Visualisation of the spectral and spatial information learned after training of EEG-ITNet for subject 3 in BCI competition IV dataset 2a in the within-subject scenario. Alpha modulated activities over the contralateral sensory and motor cortex (captured clearly by filters 4, 7, and 8 in the left panel and 4 in the middle panel) are associated with event-related desynchronisation in the alpha band during left or right hand motor imagery. Modulated alpha activity in parieto-occipital regions co-occurs in the same filters. This represents a decrease in alpha power in the contralateral hemisphere due to the need to maintain covert spatial attention toward the target limb. This pattern can be seen in other filters as well, including the first filter of the middle and right panels. The modulated activation in alpha captured over the vertex (filter 5 on the left column) likely represents alpha desynchronisation during imagination of feet movement. The beta equivalent of this event-related desynchronisation was captured by 64-sample filters 1 and 6 over the vertex and is likely also related to the imagination of foot movement. The low-frequency activations (< 5Hz) over frontal areas such as in filters 1, 3, and 6 of the left panel and filter 3 in the middle likely represent eye blinks and movements.

The spatial activation in the parieto-occipital regions in the alpha band witnessed by filters 4, 7, and 8 of the left panel may be associated with the need to maintain spatial attention toward the target limb. Studies have shown that covert spatial attention can lead to alpha desynchronisation in the contralateral hemisphere alongside an increase in alpha power ipsilaterally over the visual cortex [40-42]. So, for instance, during right-hand motor imagery, when the subjects direct their visual attention to the right visual hemifield and start imagining movement, alpha desynchronisation happens contralaterally in motor and visual cortical regions that are associated with motor imagery and visual attention, respectively. Increased gamma activity has also been reported during visual attention contralateral to the visual hemifield [36], which would explain the increase in the power spectrum between 30-40Hz in most of the filters, yet more detectable in the second filter of each inception module. Gamma-band activity can also be explained considering that the time window used for the analysis contains the period in which subjects attended to the visual cue. Please also note that visualisations >60Hz are not valid based on the Nyquist theorem as we have downsampled our signals to 125Hz.

Finally, spatial activations in low frequencies (< 5Hz) over frontal regions most likely represent eye blinks and eye movements. This pattern can be seen in filters 1, 3, and 6 of the left panel and filter 3 in the middle. While eye blinks are unlikely to play a functional role, there might be a relationship between subjects' eye movement and how they attended to and performed each motor imagery task in the experiment, which is captured as a discriminative pattern by the neural network.

## V. DISCUSSION

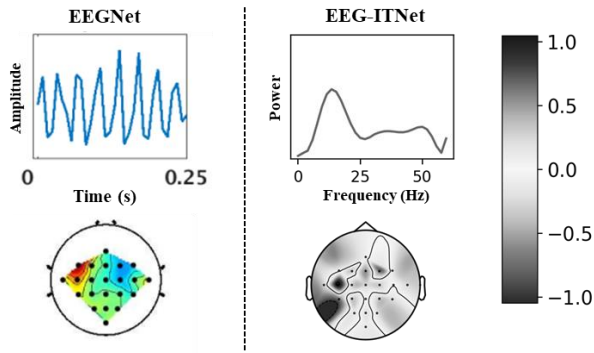

Fig. 11. Comparison of the outcome of EEGNet (left) and EEG-ITNet (right) feature visualisation techniques. In EEGNet, frequency information can be calculated based on the number of detectable cycles in the learned temporal kernel window, resulting in a single frequency component. So, for example, the above kernel from EEGNet represents 32Hz as it contains 8 cycles in 0.25 seconds.

This study proposed EEG-ITNet, an explainable CNN architecture based on inception modules and causal convolutions with dilation. Inception modules were proposed to eliminate the need to use fix-length kernels, allowing the network to learn input data patterns at different scales. In addition, causal convolutions with dilation were employed as an alternative to RNNs to learn and extract temporal dependencies and information in EEG time series. The series of depthwise causal convolutions with dilation in the form of residual blocks extracted informative discriminative features to perform the classification task in such a way that allowed EEG-ITNet to outperform well-known existing end-to-end neural networks in different scenarios.

Another significance of EEG-ITNet is its less complex structure (in terms of the number of trainable parameters) compared to other existing end-to-end architectures, including EEG-Inception and EEG-TCNet. Together with the proposed interpretable feature visualisation technique, it makes EEG-ITNet an ideal candidate for EEG analysis. Among our end-to-end comparison models, only the authors of EEGNet have attempted to explain and visualise the learned features. However, in our study, the Fourier transform combined with Savitzky-Golay filtering offered more interpretability and accuracy for network visualisation (Fig. 11). For instance, the weights of the convolutional layers applied along the time axis in EEGNet were translated as representing a single frequency component each. However, it is almost unlikely for the neural patterns of motor movements or mental tasks to be generated from a single frequency component. That is why the reader may find the smoothed power spectrum of sources generated in EEG-ITNet more intuitive and easier to interpret. Besides, the authors of EEGNet have made no attempt to explain the spatial filters from a neuroscientific viewpoint. Hence, we discussed the validity of the extracted features and supported them from a neuroscientific perspective. Our discussion was based on a fair assumption that motor imagery in our selected experiment relies on lateralised changes in the spatial attention toward the target limb and the imagination of motor execution.

## VI. CONCLUSION

We thoroughly evaluated our proposed network on two motor imagery BCI datasets in three different scenarios: within-subject, cross-subject, and cross-subject with fine-tuning. As a result, EEG-ITNet showed statistically significant improvement in performing the classification task in most of the scenarios compared to other end-to-end architectures, including EEG-Inception, EEGNet, and EEG-TCNet. Furthermore, we introduced a visualisation technique based on Fourier transform and Savitzky-Golay filtering to explain the network learning criteria. Although initially showed promising results, the performance of EEG-ITNet may be improved even further by finding the optimum hyperparameters for each scenario. In addition, finding a way to visualise and explain temporal features and dependencies learned inside our TC block would be of great importance. This can further improve the interpretability of EEG-ITNet and help the neuroscientific community with implementing deep neural networks such as ours in their research. Besides, we are interested in further investigating the effect of network deepness on its performance. These will be our focus in future studies. Nevertheless, we believe that the concepts introduced in this study can help to develop more robust, interpretable, and high-accuracy BCI systems.

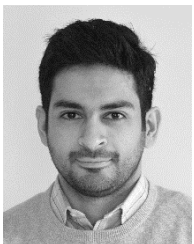

**Abbas Salami** received the B.Sc. degree in Electrical Engineering from Ferdowsi University Mashhad, Iran, in 2015, and the M.Sc. degree in Biomedical Engineering from Amirkabir University of Technology, Iran, in 2018. He is currently pursuing a PhD degree in Computer Science at the University of Essex, UK. His current research interests include brain-computer interfaces, biological signals processing, and deep learning.

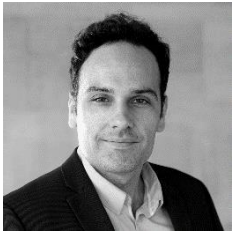

**Javier Andreu-Perez** (PhD '12, SMIEEE' 19) is Senior Lecturer and Chair of the Smart Health Technologies Group within the Centre for Computational Intelligence, University of Essex (UK). His main expertise is in employing and developing novel Artificial Intelligence methods for neuro/bio-engineering and health informatics. Javier's research interest is in deep learning, fuzzy systems, and human-centered artificial intelligence, particularly developing novel methods focusing on trustworthy and explainable AI to work along with humans. Javier has published his research in prestigious journals from IEEE, Nature, Springer and Elsevier publishers, mostly in top artificial intelligence and neuroscience journals. Javier's work in Artificial Intelligence and Biomedical engineering has attracted nearly 3000 citations. Javier also contributes as associate Editor-in-Chief of the Journal Neurocomputing (Elsevier) and other editorials from other prestigious journals in computational intelligence and emerging technologies. Javier has also received several personal research awards (fellowships) and has been the primary investigator of projects funded by the United Kingdom research councils, international organizations, and corporations.

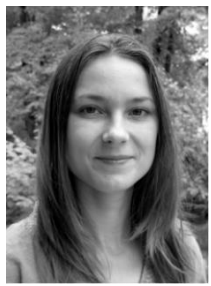

**Helge Gillmeister** received a BSc in Psychology with Cognitive Science from UCL (University of London, UK) in 2000 and a PhD in Psychology from Birkbeck (University of London, UK) in 2005. She is currently a Reader at the Department of Psychology and Centre for Brain Science at the University of Essex (UK). She has previously worked at the MRC Cognition and Brain Sciences Unit (Cambridge, UK), Birkbeck, UCL and City (University of London, UK). Her research in cognitive neuroscience is centred around the bodily self and its sensorimotor foundations in health and disease. Dr Gillmeister is the communications officer for the British Association for Cognitive Neuroscience, and also a member of the Experimental Psychology Society, the European Brain and Behaviour Society and the European Society for Cognitive and Affective Neuroscience.